\documentclass[switch,subdepth]{article} 
\usepackage{iclr2023_conference,times}
\usepackage{graphicx}

\usepackage{amsmath,amsfonts,bm}

\def\eqref#1{equation~\ref{#1}}

\def\1{\bm{1}}

\def\vh{{\bm{h}}}

\def\vx{{\bm{x}}}
\def\vy{{\bm{y}}}

\DeclareMathAlphabet{\mathsfit}{\encodingdefault}{\sfdefault}{m}{sl}
\SetMathAlphabet{\mathsfit}{bold}{\encodingdefault}{\sfdefault}{bx}{n}

\usepackage{hyperref}
\usepackage{url}
\usepackage{xspace}

\usepackage{enumitem}
\usepackage{multirow}
\usepackage{booktabs}
\usepackage{arydshln}
\usepackage{adjustbox}
\usepackage{amsmath}
\usepackage{array}
\usepackage{makecell}
\usepackage{fourier} 
\usepackage{listings,lstautogobble}
\usepackage{fancyvrb}
\usepackage{fvextra}
\usepackage{caption}
\usepackage{graphicx}
\usepackage{verbatim}
\usepackage{float}

\newcommand{\vmytheta}{\boldsymbol{\theta}}
\usepackage[most]{tcolorbox}
\usepackage{alltt}
\usepackage{lineno}
\newtcolorbox{AIbox}[2][]{aibox,title=#2,#1}
\tcbset{
  aibox/.style={
    top=10pt,
    colframe=black,
    colbacktitle=black,
    enhanced,
    center,
    attach boxed title to top left={yshift=-0.1in,xshift=0.15in},
    boxed title style={boxrule=0pt,colframe=white,},
  }
}

\definecolor{forestgreen}{rgb}{0.13, 0.55, 0.13}

\newcommand{\model}[1]{\textsc{#1}\xspace}
\newcommand{\ours}{\model{Macaw-LLM}}

\title{\ours: Multi-Modal Language Modeling with Image, Audio, Video, and Text Integration}

\iclrfinalcopy

\author{Chenyang Lyu$^{1,2}$, Minghao Wu$^{3}$, Longyue Wang$^{1}$\thanks{Longyue Wang is the corresponding author: \texttt{vinnlywang@tencent.com}.}~~, Xinting Huang$^{1}$, \\\bf Bingshuai Liu$^{1}$, \bf Zefeng Du$^{1}$, Shuming Shi$^{1}$ \& Zhaopeng Tu$^{1}$ \\
$^{1}$Tencent AI Lab \qquad\qquad
$^{2}$Dublin City University \qquad\qquad
$^{3}$Monash University\\
\texttt{chenyang.lyu2@mail.dcu.ie, minghao.wu@monash.edu},\\
\texttt{\{timxthuang,bsliu,zefengdu,shumingshi,zptu\}@tencent.com}}

\newcommand{\vQ}{\pmb{Q}}
\newcommand{\vK}{\pmb{K}}
\newcommand{\vV}{\pmb{V}}
\newcommand{\vE}{\pmb{E}}

\begin{document}

\renewcommand{\tableautorefname}{Table}
\renewcommand{\sectionautorefname}{Section}
\renewcommand{\subsectionautorefname}{Section}
\renewcommand{\subsubsectionautorefname}{Section}
\renewcommand{\figureautorefname}{Figure}
\renewcommand{\equationautorefname}{Equation}
\newcommand{\linenoautorefname}{Line}

\maketitle

\begin{abstract}
Although instruction-tuned large language models (LLMs) have exhibited remarkable capabilities across various NLP tasks, their effectiveness on other data modalities beyond text has not been fully studied. 
In this work, we propose \ours, a novel multi-modal LLM that seamlessly integrates visual, audio, and textual information. 
\ours consists of three main components: a modality module for encoding multi-modal data, a cognitive module for harnessing pretrained LLMs, and an {\em alignment module} for harmonizing diverse representations. 
Our novel alignment module seamlessly bridges multi-modal features to textual features, simplifying the adaptation process from the modality modules to the congitive module. 
In addition, we construct a large-scale multi-modal instruction dataset in terms of multi-turn dialogue, including 69K image instances and 50K video instances. 
We have made our data, code and model publicly available, which we hope can pave the way for future research in multi-modal LLMs and expand the capabilities of LLMs to handle diverse data modalities and address complex real-world scenarios.
\end{abstract}

\begin{center}
\includegraphics[height=45pt]{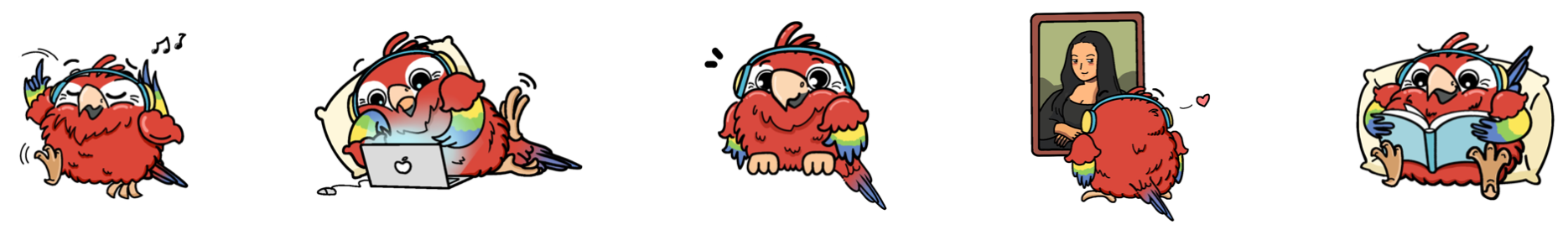}
{\color{blue}\url{https://github.com/lyuchenyang/Macaw-LLM}}
\end{center}

\section{Introduction}

Instruction-tuned large language models (LLMs) have demonstrated impressive capabilities across various domains, exhibiting zero-shot generalization without the need for task-specific fine-tuning \citep{ouyang2022training, wei2022finetuned, DBLP:conf/iclr/SanhWRBSACSRDBX22, DBLP:journals/corr/abs-2210-11416, DBLP:journals/corr/abs-2303-08774}.
However, these models are primarily limited to processing text-based data.
Previous research on multi-modal pre-training has shown promise in aligning knowledge from different modalities within a shared latent space \citep{DBLP:conf/icml/WangYMLBLMZZY22, DBLP:conf/nips/AlayracDLMBHLMM22, bao2022beit, DBLP:journals/corr/abs-2208-10442}.
Furthermore, there is a recent line of research papers focusing on enabling multi-modal pre-trained models to understand and follow instructions \citep{DBLP:journals/corr/abs-2212-10773, DBLP:journals/corr/abs-2304-10592, liu2023visual_llava, DBLP:journals/corr/abs-2305-03726,DBLP:journals/corr/abs-2305-04790,dai2023instructblip, pandagpt,huang2023language_kosmos_1}.

In this work, we propose \ours, a multi-modal instruction-tuned LLM that integrates four different modalities, including image, video, audio, and text, into one single model. We propose a novel alignment approach that aligns multi-modal features to the embeddings of LLMs, which produces aligned features that are closer to the textual features of language models and can be naturally injected into the input sequence of LLMs. A key motivation for our approach is to streamline the adaptation process for LLMs. In particular, \ours employs a one-stage instruction fine-tuning process, promoting a simpler learning experience. Previous multi-modal systems typically require two-stage training~\cite{DBLP:journals/corr/abs-2301-12597,DBLP:journals/corr/abs-2304-10592,liu2023visual_llava,dai2023instructblip}, where the first stage usually trains the projection layer for alignment between multi-modal features and text features, and the second stage is the general instruction fine-tuning for LLMs. In contrast, our approach aligns the multi-modal features to the embedding layer of LLMs, which produce aligned features based on LLMs embeddings that can be naturally injected into the input sequence of LLMs. This makes our approach more advantageous.

To address the limitations of current multi-modal datasets that predominantly emphasize specific task types, we create our \ours instruction dataset, which is described in \autoref{sec:dataset}. This dataset covers a wide range of instructional tasks and combines various data modalities, making it more diverse and better-suited for multi-modal instruction-tuned LLMs. We utilize the remarkable generative capability of current LLMs, such as \model{GPT-3.5-Turbo}, to curate this dataset, ensuring the target text properly aligns with human instructions.

Our contributions in this work can be summarized as follows:
\begin{itemize}[leftmargin=*,topsep=0.1em,itemsep=0.1em,parsep=0.1em]
    \item We propose a novel architecture for multi-modal language modeling, which jointly learns to {\bf align} multi-modal features and textual features and {\bf generate} output sequence.
    
    \item We release \ours instruction dataset, a large-scale {\bf multi-modal instruction dataset} that covers diverse instructional tasks leveraging image and video modalities, which facilitates future work on multi-modal LLMs. 

\end{itemize}
\section{Related Work}

\paragraph{Instruction-Tuned Large Language Models}
Large language models (LLMs) have showcased exceptional generative capabilities in a wide range of natural language processing (NLP) tasks \citep{NEURIPS2020_1457c0d6, DBLP:journals/corr/abs-2201-08239, DBLP:journals/corr/abs-2203-15556, DBLP:journals/corr/abs-2204-02311}.
By leveraging techniques such as supervised instruction tuning and reinforcement learning from human feedback (RLHF), LLMs exhibit remarkable few- and zero-shot generalization capabilities \citep{ouyang2022training, wei2022finetuned, DBLP:conf/iclr/SanhWRBSACSRDBX22, DBLP:journals/corr/abs-2210-11416, DBLP:journals/corr/abs-2211-01786, DBLP:journals/corr/abs-2303-08774, DBLP:journals/corr/abs-2305-10403}.
Recently, \citet{DBLP:journals/corr/abs-2212-10560} highlight the lack of diversity in human-written instructions and demonstrate that machine-generated instructions can be used for instruction tuning. Since then, several instruction-tuned LLMs have been fine-tuned using various machine-generated instruction datasets \citep{alpaca, vicuna2023, li2023bactrian}.
More surprisingly, \citet{DBLP:journals/corr/abs-2304-14402} reveal that instruction-following is not solely a property of LLMs, as even relatively small language models can follow instructions when fine-tuned on large-scale instruction datasets.

\paragraph{Multi-Modality}
Drawing inspiration from the human learning process, artificial intelligence (AI) researchers are actively exploring the combination of different modalities to train deep learning models.
With the success of LLMs, feature alignment among multiple modalities has attracted great interest for its applications. 
There is a line of research works that learns a joint embedding space for multiple modalities \citep{DBLP:conf/icml/RadfordKHRGASAM21, DBLP:conf/icml/BaevskiHXBGA22, DBLP:journals/corr/abs-2305-05665}.
Some researches also attempt to combine the pre-trained vision-only and language-only models, showcasing impressive zero-shot capabilities \citep{DBLP:conf/nips/AlayracDLMBHLMM22, DBLP:journals/corr/abs-2301-12597, DBLP:journals/corr/abs-2205-02655}.
More recently, a number of works explore to enable the multi-modal LLMs to follow the instructions \citep{DBLP:journals/corr/abs-2304-10592, DBLP:journals/corr/abs-2304-14178, DBLP:journals/corr/abs-2305-03726, DBLP:journals/corr/abs-2305-04160, DBLP:journals/corr/abs-2305-04790, dai2023instructblip}.
\citet{DBLP:journals/corr/abs-2212-10773} introduce MultiInstruct, the first multi-modal instruction tuning benchmark dataset covering a wide range of multi-modal tasks and categories.
\citet{liu2023visual_llava} explore the multi-modal instruction-tuning using the machine-generated data.
\citet{pandagpt} allow the textual LLMs to support six modalities using the parameter-efficient fine-tuning technique LoRA.

\paragraph{Our Work}
In this work, we propose \ours, a multi-modal LLM that effectively integrates information from visual, audio, and textual modalities, enabling it to comprehend and execute instructions accurately.

\section{Methodology}


In this section, we provide a comprehensive description of \ours. 
We begin by presenting an outline of the model architecture, followed by a detailed description of each individual module within \ours, namely the modality module, alignment module, and cognitive module. 
Lastly, we provide an in-depth explanation of the training process of \ours.

\begin{figure}[t]
    \centering
    \includegraphics[width=0.85\textwidth]{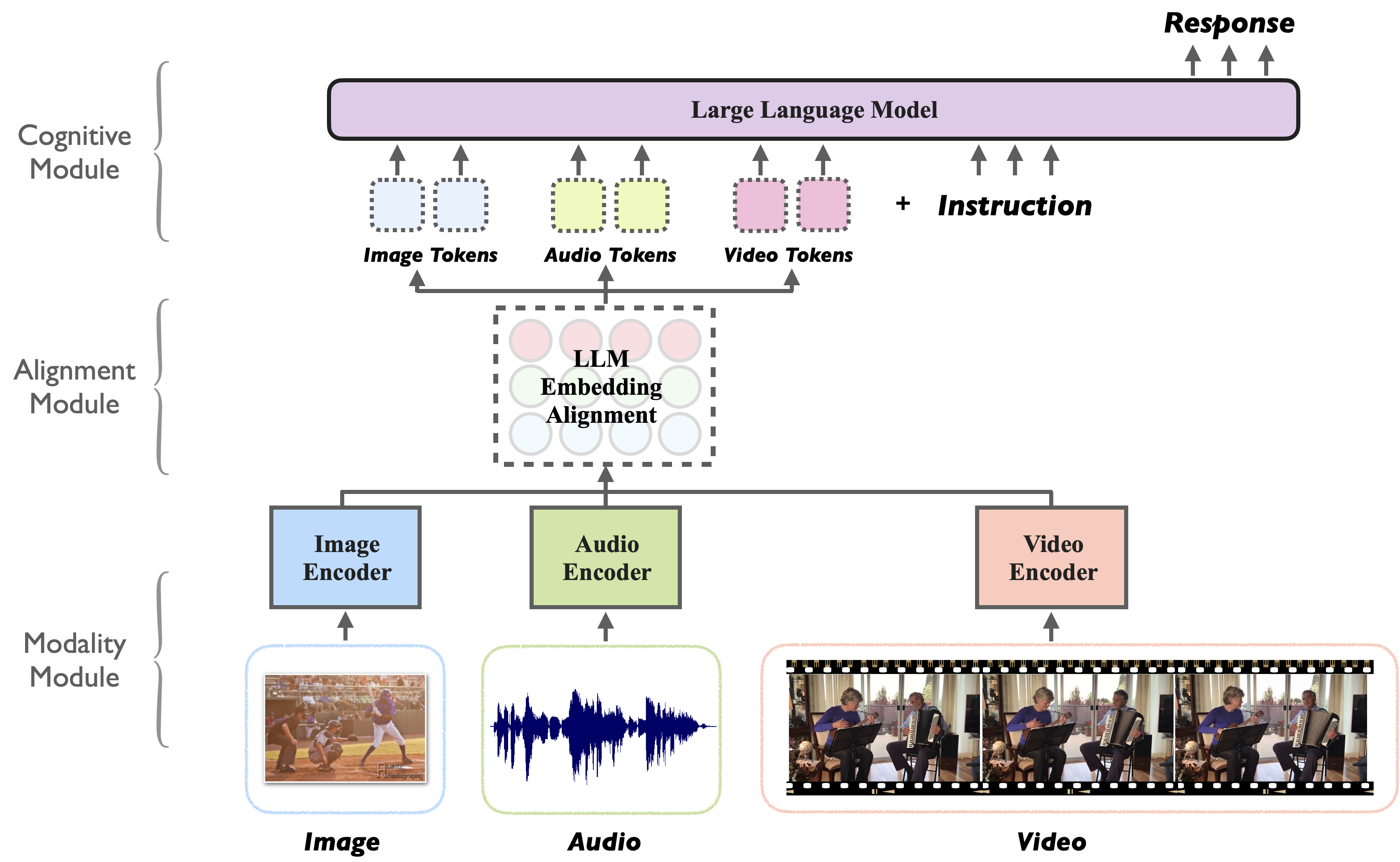}
    \caption{An overview of \ours model architecture.}
    \label{fig:architecture}
\end{figure}

\subsection{Model Overview}

We present an overview of \ours in this section.
As shown in \autoref{fig:architecture}, there are three major modules in \ours as follows:
\begin{itemize}[leftmargin=*,topsep=0.1em,itemsep=0.1em,parsep=0.1em]
    \item \textbf{Modality Module}: Existing LLMs primarily focus on processing textual information. To incorporate additional modalities such as visual and audio data, we integrate extra modality encoders into \ours. This enhancement enables our \ours to handle multiple modalities effectively.
    \item \textbf{Alignment Module}: Since each modality encoder is trained independently, the learned representations of different modalities may not be directly compatible. To address this, we propose the alignment module, which unifies the representations from different modalities, enabling effective integration of multi-modal information.
    \item \textbf{Cognitive Module}:  LLMs have demonstrated remarkable capability in understanding and following human instructions. In \ours, we leverage pretrained LLMs as our cognitive module, which forms the foundation of \ours. It is worth noting that the cognitive module also serves as the textual modality encoder in our approach.
\end{itemize}

\autoref{fig:architecture} provides a visual representation of the \ours architecture, while \autoref{sec:modality} and \autoref{sec:alignment} offer detailed explanations of the modality module and alignment module, respectively.
As the cognitive module of \ours, the effectiveness of instruction-tuned LLMs has been demonstrated by several previous works \citep{ouyang2022training, wei2022finetuned, DBLP:journals/corr/abs-2303-08774, alpaca, vicuna2023, DBLP:journals/corr/abs-2305-10403}, and we follow their practices in our \ours.

\subsection{Modality Module}
\label{sec:modality}
Existing LLMs are highly powerful but typically limited to processing only textual information. 
In this section, we describe how we encode information from different modalities.

\paragraph{Visual Modality Encoder}
\citet{DBLP:conf/icml/RadfordKHRGASAM21} propose a novel framework, known as \model{CLIP} \citep{DBLP:conf/icml/RadfordKHRGASAM21}, which exploits a significantly wider range of supervision by directly learning from unprocessed textual data related to images. In this work, we utilize the capabilities of \model{CLIP-ViT-B/16} for encoding visual information, including images and video frames.

\paragraph{Audio Modality Encoder}
\citet{DBLP:journals/corr/abs-2212-04356} introduce a novel multilingual speech recognition model called \model{Whisper} \citep{DBLP:journals/corr/abs-2212-04356}. This model is trained on a vast audio dataset with weak supervision. In \ours, we leverage the power of \model{Whisper-base} to encode the audio signals, thereby extracting meaningful representations from the audio data.

\paragraph{Textual Modality Encoder}
LLMs are commonly pre-trained on the massive text corpora, so instruction-tuned LLMs can naturally process text information. In this work, we consider \model{LLaMA-7B} \citep{DBLP:journals/corr/abs-2302-13971} as the foundation of \ours.

We acknowledge the existence of numerous publicly available pre-trained models that can serve as modality encoders. However, we leave the investigation of their utility to the future work.

\subsection{Alignment Module}
\label{sec:alignment}
Modality encoders are typically trained separately, leading to potential discrepancies in the representations generated by different encoders. As a result, it becomes crucial to align these independent representations within a joint space. In this section, we outline the approach we employ to align these representations.

\paragraph{Multi-Head Self-Attention (MHSA)}
Scaled dot-product attention is a fundamental component of the Transformer model \citep{transformer}. It operates on three inputs: the query vector $\vQ \in \mathbb{R}^{n_q \times d_k}$, the key vector $\vK \in \mathbb{R}^{n_k \times d_k}$, and the value vector $\vV \in \mathbb{R}^{n_k \times d_v}$. This attention mechanism calculates attention weights by comparing the queries $\vQ$ with the keys $\vK$. It then uses these weights to update the query representations through a weighted sum of the values $\vV$ and can be described as follows:
\begin{align}
    \label{eq:attn}
    \textrm{Attn}(\vQ, \vK, \vV) = \textrm{softmax}(\frac{\vQ\vK^{\top}}{\sqrt{d_k}}) \vV,
\end{align}
where $d_k$ is the dimensionality of the key and query vectors, and $n_q$ and $n_k$ are the number of queries and keys, respectively.

\paragraph{Modality Alignment}

The alignment strategy is designed to efficiently connect multi-modal features with textual features, facilitating quicker adaptation. In this work, we refer to the image and video features obtained from our visual modality encoder (i.e. \model{CLIP}) as $\vx_{i} \in \mathbb{R}^{L_i \times d_i}$ and $\vx_{v} \in \mathbb{R}^{L_v \times d_v}$, respectively. Additionally, we denote the audio features from the audio modality encoder (i.e. \model{Whisper}) as $\vx_a \in \mathbb{R}^{L_a \times d_a}$. The process of modality alignment is outlined as follows:


\begin{enumerate}[leftmargin=*,topsep=0.1em,itemsep=0.1em,parsep=0.1em]
\item \textbf{Encoding}: We firstly leverage the pre-trained models ,\model{CLIP} and \model{Whisper}, to encode multi-modal features:
\begin{equation}
    \vh_i = \text{\model{CLIP}}(\vx_i), \quad  \vh_v = \text{\model{CLIP}}(\vx_v), \quad \vh_a = \text{\model{Whisper}}(\vx_a),
\end{equation}
where $\vh_{i} \in \mathbb{R}^{L_i \times d_h}$, $\vh_{v} \in \mathbb{R}^{L_v \times d_h}$ and $\vh_{a} \in \mathbb{R}^{L_a \times d_h}$ are image, video, and audio features, respectively, and $d_h$ is the dimension of modality-specific features.

\item \textbf{Transformation}: To reduce computational costs and minimize the number of tokens in the prefix, we employ a 1-D convolutional layer to compress the length of the multi-modal features to a smaller and fixed value. Subsequently, a linear layer is employed to adjust the hidden size of the features, aligning it with the size of the LLMs embeddings as follows:
\begin{equation}
\label{eq:transform}
    \vh_i' = \text{Linear}(\text{Conv1D}(\vh_i)), \quad \vh_v' = \text{Linear}(\text{Conv1D}(\vh_v)), \quad \vh_a' = \text{Linear}(\text{Conv1D}(\vh_a)),
\end{equation}
where $h_i' \in \mathbb{R}^{L' \times d_e}$, $h_v' \in \mathbb{R}^{L' \times d_e}$, and $h_a' \in \mathbb{R}^{L' \times d_e}$ are the transformed features with a fixed length of $L'$ and an embedding dimension of $d_e$. The value of $L'$ is significantly smaller than $L_i$, $L_v$, and $L_a$, while $d_e$ corresponds to the dimensionality of the embedding matrix $\vE \in \mathbb{R}^{V \times d_e}$ associated with the textual LLMs (i.e. \model{LLaMA} in this work).

\item \textbf{Alignment}: 
Each modality encoder is trained separately, resulting in distinct representations for different modalities. To establish a common representation space, it becomes necessary to align these representations across modalities.
In this work, we consider the transformed visual and audio modality representations obtained in \autoref{eq:transform} as the \textit{soft tokens} of LLM, the cognitive model, so we propose to align the visual and audio representations with the textual embedding space using the attention mechanism in \autoref{eq:attn} as follows:
\begin{align}
    \label{eq:align}
    \vh^{a} = \textrm{Attn}(\vh', \vE, \vE),
\end{align}
where $\vh'$ is the modality representation obtained in \autoref{eq:transform} (\i.e. $\vh_i'$, $\vh_v'$, and $\vh_a'$) and $\vh^{a}$ is the corresponding aligned representation, specifically, $\vh^{a}_{i}$, $\vh^{a}_{v}$, and $\vh^{a}_{a}$.
After such an alignment operation facilitated by the attention mechanism, the LLM (cognitive module) can seamlessly process the representations from various modalities.

\item \textbf{Integration}: 
The integration of aligned modality representations into the instruction can be achieved effortlessly through the concatenation operation. Given the aligned modality representations, the integration can be defined as follows: 
\begin{align}
    \label{eq:concat}
    \vx = [\vh^{a}_{i}:\vh^{a}_{v}:\vh^{a}_{a}:\textrm{Embed}(\vx_{\textrm{t}})],
\end{align}
where $[:]$ represents the concatenation operation, $\vx$ represents the multi-modal instruction, $\vx_{\textrm{t}}$ represents the sequence of tokens in the textual instruction, and $\textrm{Embed}(\vx_{\textrm{t}})$ represents the sequence of embeddings of $\vx_{\textrm{t}}$.

\end{enumerate}


In this section, we describe how we align the multi-modality representation into a shared representation space using the attention mechanism.
It is important to note that our model, \ours, has the capability to process multiple modalities concurrently, while the textual instruction $\vx_{\textrm{t}}$ is always necessary as part of the instruction $\vx$. We intend to investigate the direct utilization of visual or audio instructions in our future work.

\subsection{One-Step Instruction Fine-Tuning}
The common multi-modal practice in previous works involves two-step training \citep{DBLP:journals/corr/abs-2301-12597,liu2023visual_llava,dai2023instructblip}. The first step focuses on training the projection layer to align multi-modal features with textual features, while the second step involves fine-tuning the general instruction for LLMs.
In contrast, our approach, \ours, simplifies the adaptation process by employing a one-step instruction fine-tuning approach. This approach ensures coherent alignment across the modalities and eliminates the potential risk of error propagation that can occur in multi-step fine-tuning procedures.



In this work, we fine-tune all the parameters $\vmytheta$ in \ours, and the objective is to minimize the negative log-likelihood over the response $\vy$ with respect to $\vmytheta$ as follows:
\begin{equation}
    \mathcal{L}(\vy; \vmytheta) = -\sum_{j=1}^{N} \log P(y_j | \vx; \vmytheta),
\end{equation}
where $N$ denotes the number of tokens in $\vy$ and $y_j$ is the $j$-th token in $\vy$.
By employing such a one-step fine-tuning strategy, \ours can effectively harmonize the different modules.



\begin{figure}[t]
    \centering
    \includegraphics[width=1\textwidth]{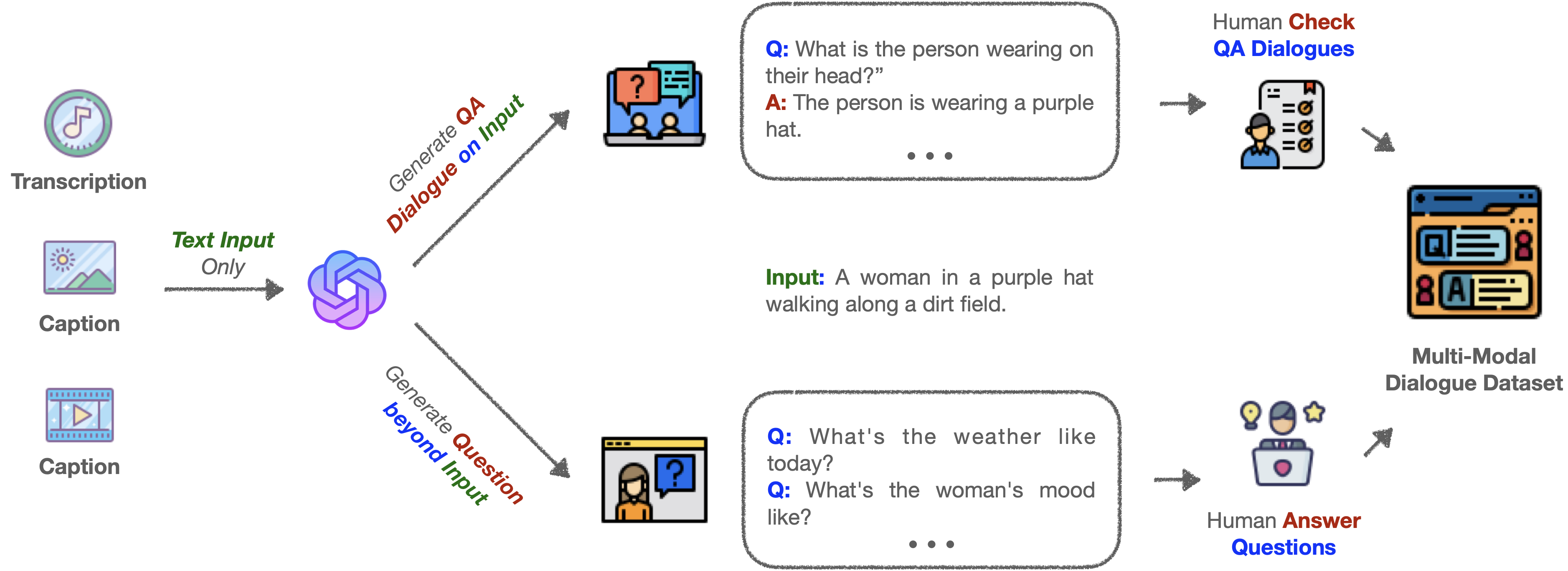}
    \caption{The process of constructing a multi-modal instruction corpus. GPT-4 is prompted to generate instruction-response examples from text input, which then correspond to its multi-modal contents. Human verification and annotation are incorporated to ensure the accuracy.}
    \label{fig:dataset_creation}
\end{figure}

\begin{figure}[t]

\begin{AIbox}{Building Multi-Modal Dialogue with GPT-4}
\small{\bf Prompt:} \\

This is the {\color{forestgreen} \bf caption of an image/video: \{$\dots$\}}. This image/video contains important information that needs to be conveyed through high-quality instructions. \\

Your task is to provide ten pairs of instructions and responses that are related to the content of the image/video caption like dialogue concentrating on the content of the image/video without explicitly mentioning the caption or the word 'caption'. \\

Your focus should be on describing, explaining, or analyzing various aspects of the image/video, as well as providing some QA pairs. The purpose of this exercise is to fine-tune a language model so that it can generate accurate and relevant responses. \\
    
In each pair, the first line should start with "Q:" and contain an instruction related to the image/video, while the second line should start with "A:" and provide a response to the instruction. \\
    
Please ensure that your instructions are diverse and of high quality, accurately reflecting the content of the image and providing useful information to the language model:
\hfill
\tcbline
{\bf Output}:\\

\textcolor{blue}{\bf Q}: {\em Can you describe the color of the river in the image? }

\textcolor{red}{\bf A}: {\bf The river in the image appears to be a tranquil shade of blue.} \\

\textcolor{blue}{\bf Q}: {\em What type of boat is the man in the image paddling?}

\textcolor{red}{\bf A}: {\bf The man in the image is paddling a kayak.} \\

\textcolor{blue}{\bf Q}: {\em How do you think the man in the image is feeling while paddling down the river?} 

\textcolor{red}{\bf A}: {\bf Judging by the peaceful surroundings and the calm pace of the paddling, it's likely that the man in the image is feeling relaxed and at ease.} \\
......

\end{AIbox}

\caption{Our instruction generation prompt for creating instruction-response pairs related to the content of an image/video caption using GPT-4. The objective is to create high-quality instructions and responses without explicitly mentioning the caption, aiming to improve the language model's ability to generate accurate and relevant responses.}	
\label{fig:prompt_example}
\end{figure}

\section{\ours Instruction Dataset}
\label{sec:dataset}
Current multi-modal datasets, such as visual question answering \citep{Antol_2015_ICCV, Goyal_2017_CVPR}, summarization \citep{li-etal-2017-multi, 10.1145/3584700}, and dialogue \citep{shuster-etal-2021-multi, sun-etal-2022-multimodal}, predominantly emphasize specific task types, resulting in a limited diversity of tasks.
Additionally, the target text in these datasets often lacks proper alignment with the style of human-written text, making it difficult for models fine-tuned on such data to effectively follow human instructions.
To address these limitations, we utilize the remarkable generative capability of current LLMs (i.e. \model{GPT-3.5-Turbo}) to curate our \ours instruction dataset.


To generate the dataset, we utilize the power of \model{GPT-3.5-Turbo}. We provide it with a prompt in the form of an image or video caption (see \autoref{fig:prompt_example}). To optimize the generation process and improve efficiency, we generate 10 instruction-response pairs within a single query.
For image caption data, we rely on the MS COCO dataset \citep{lin2014microsoft_coco}. It consists of 328,000 images accompanied by captions. From this dataset, we randomly select a subset of 10,000 images with their respective captions to create our dataset.
In addition to image data, we incorporate video caption data from two datasets: Charades \citep{DBLP:journals/corr/SigurdssonRFLG16_charades} and AVSD \citep{alamri2019audiovisual_avsd}. These datasets collectively contain 9,848 videos with captions, which we utilize to create our own dataset.

We repeat this process and obtain approximately 69K examples based on COCO image captions and about 50K examples based on Charades and AVSD video captions. The dataset creation process is illustrated in \autoref{fig:dataset_creation}. \autoref{tab:data} provides statistics about the dataset, including the number of items, the word count of instructions and responses, and examples of each type.

Our current dataset is focused on single-turn dialogues, but we acknowledge the significance of including multi-turn dialogues and expanding the dataset to encompass a wider range of multi-modal content. To address this, we are actively engaged in the process of incorporating multi-turn dialogues and diversifying the dataset to enhance its richness. These additions will greatly contribute to enriching the dataset and will be beneficial for the fine-tuning process of language learning models (LLMs).

\begin{table}[t]
\centering
\begin{tabular}{p{1.5cm}cccp{6cm}}
\toprule
\textbf{Dataset} & \textbf{Items} & \textbf{Ins. Len.} & \textbf{Res. Len.} & \textbf{Example} \\ 
\midrule
COCO & 69,314 & 10.1 & 15.7 &  
\begin{minipage}{0.44\textwidth}
  \centering
  \includegraphics[width=.4\linewidth]{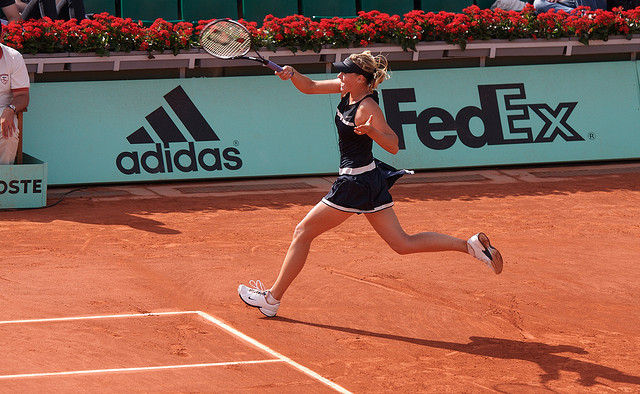}
  \par
  \raggedright
  {\bf Caption}: A woman is trying to hit a ball with a tennis racket. \\
  \vspace{3pt}
  {\bf Instruction}: What sport is the woman playing in the image? \\
    \vspace{3pt}
  {\bf Response}: The woman in the image is playing tennis with a racket.
\end{minipage} 
\\ 
\midrule
Charades/ AVSD & 50,656 & 10.2 & 14.9 & 

\begin{minipage}{0.44\textwidth}
  \centering
  \includegraphics[width=.4\linewidth]{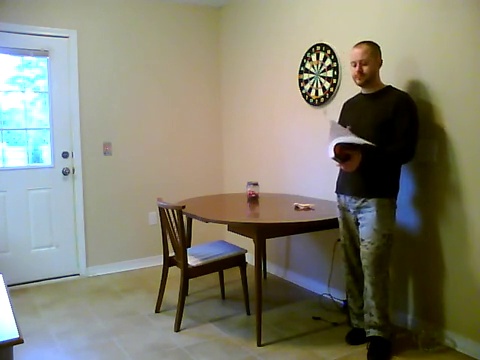}
  \par
  \raggedright
  {\bf Caption}: A person is standing next to a table holding a sandwich in their hand. They sit the sandwich on the table and pick up a book. \\
    \vspace{3pt}
  {\bf Instruction}: What does the person do right after setting down the sandwich? \\
    \vspace{3pt}
  {\bf Response}: The person picks up a book that is on the table.
\end{minipage} 
\\ 
\bottomrule
\end{tabular}
\caption{Statistics and examples of our instruction dataset. We report the number of items, average instruction length, and average response length of our instruction dataset, based on COCO, Charades, and AVSD. }
\label{tab:data}
\end{table}
\section{Experimental Setup}

\subsection{Dataset}
In this study, we utilize instruction data from three different sources:

\begin{itemize}[leftmargin=*,topsep=0.1em,itemsep=0.1em,parsep=0.1em]
    \item \textbf{Text instruction dataset}: For textual instruction-tuning, we make use of the Alpaca instruction dataset \citep{alpaca}, which comprises approximately 52,000 instruction-response examples distilled from the \model{Text-Davinci-003} model.

    \item \textbf{Image instruction dataset}: To create an image instruction dataset, we curate around 69K instruction-response pairs by generating them from COCO image captions \citep{lin2014microsoft_coco} using \model{GPT-3.5-Turbo} as described in \autoref{sec:dataset}.

    \item \textbf{Video instruction data}: We generate approximately 50K video instruction-response examples by utilizing the video captions from the Charades \citep{DBLP:journals/corr/SigurdssonRFLG16_charades} and AVSD \citep{alamri2019audiovisual_avsd} datasets using \model{GPT-3.5-Turbo} as described in \autoref{sec:dataset}.
\end{itemize}
In practice, we randomly sample 50K examples from each type of instruction data and combine them to form a final training dataset consisting of 150K examples. Note that the audio inputs are currently associated with the video instruction data and we are actively in the process of creating the audio instruction dataset.

\subsection{Hyperparameters}
We utilize DeepSpeed \citep{DBLP:conf/kdd/RasleyRRH20_deepspeed} for optimization during the training process. The training is conducted on 8 Nvidia A100 GPUs. For each device, the training batch size is set to 4. We employ a gradient accumulation step of 3.  The model is trained for 5 epochs, with a learning rate of $3\times10^{-5}$. The warmup ratio is 0.03, along with a cosine learning rate scheduler. The maximum sequence length is fixed at 512. We use FP16 precision for both training and inference.

\section{Examples}

To showcase the effectiveness and potential of our proposed \ours in creating human-like conversational agents, this section provides compelling examples that demonstrate the system's remarkable ability to understand and generate responses related to visual content.  These examples vividly illustrate how \ours seamlessly processes and integrates multiple modalities of information, such as visuals and audio, within the domain of natural language processing (NLP). By generating informative, relevant, and coherent responses to a wide range of questions, \ours clearly demonstrates its proficiency in NLP and underscores its potential for developing highly effective human-machine communication interfaces.

\begin{figure}[ht]
    \centering
    \includegraphics[width=0.75\textwidth]{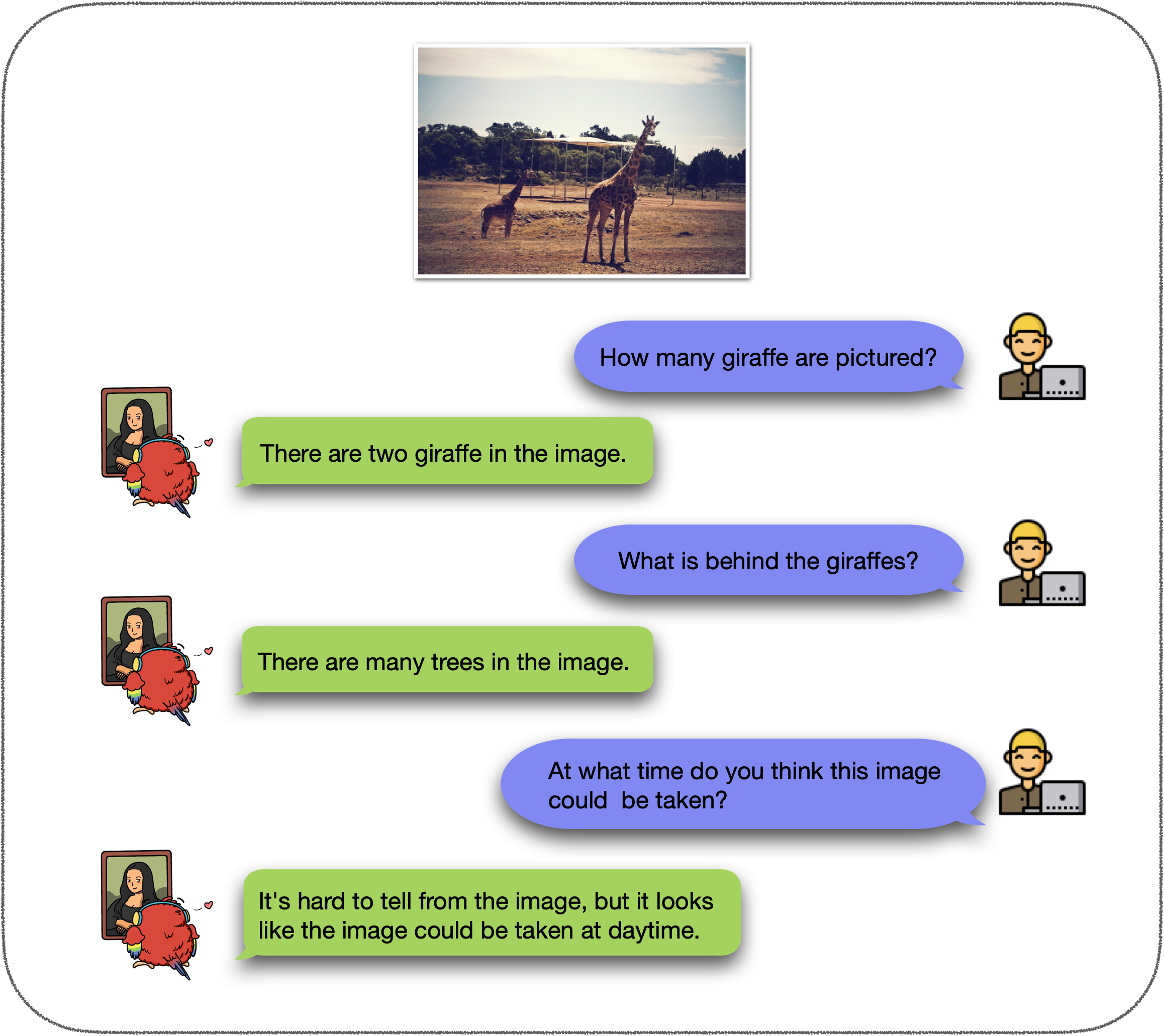}
    \caption{
    An example showcasing \ours's basic capability in {\bf image-grounded question answering}. The image features two giraffes with a backdrop of numerous trees. \ours can identifies these contents and infers that the photo was taken at daytime.
}
    \label{fig:example_1}
\end{figure}

\begin{figure}[ht]
    \centering
    \includegraphics[width=0.75\textwidth]{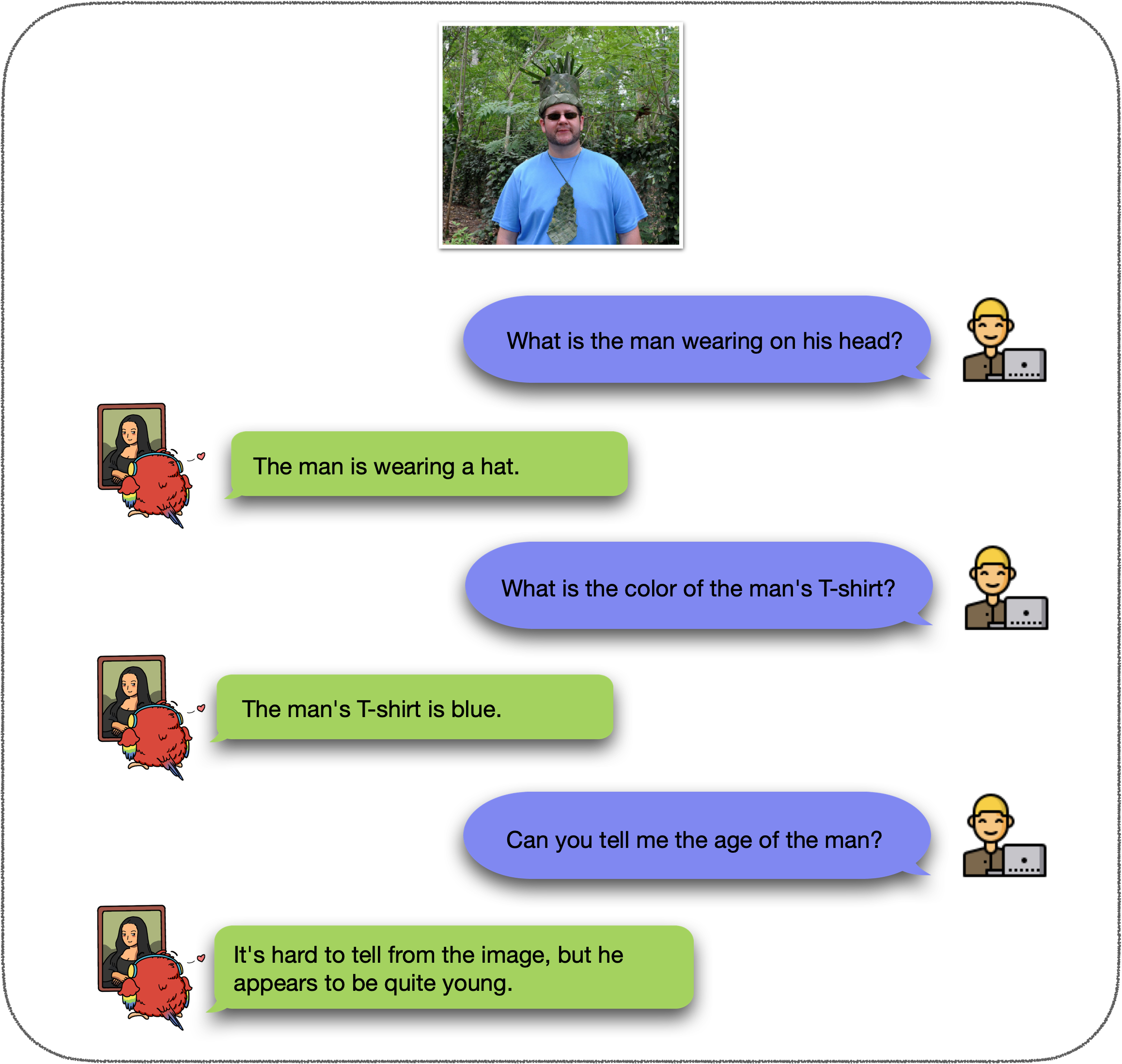}
    \caption{
    An example showcasing \ours's capability in {\bf image-grounded understanding and reasoning}. As seen, \ours can comprehend fundamental objects, such as a hat and a T-shirt. Besides, it tries to estimate the age of the man.}
    \label{fig:example_2}
\end{figure}

\begin{figure}[ht]
    \centering
    \includegraphics[width=0.75\textwidth]{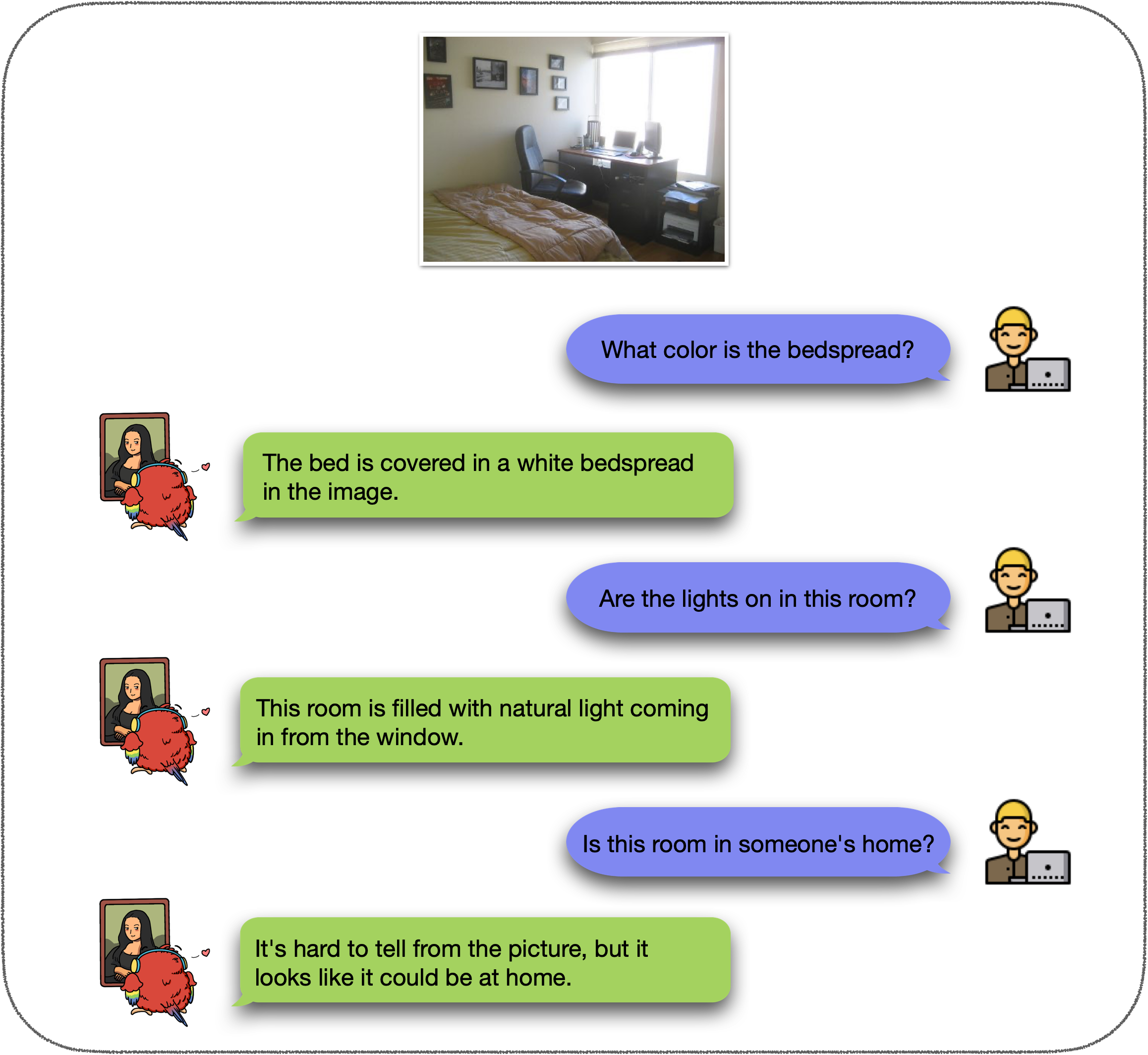}
    \caption{
    An example showcasing \ours's capability on {\bf recognizing color and light}. Besides, \ours estimate the location of the room.
    }
    \label{fig:example_3}
\end{figure}

\begin{figure}[ht]
    \centering
    \includegraphics[width=0.75\textwidth]{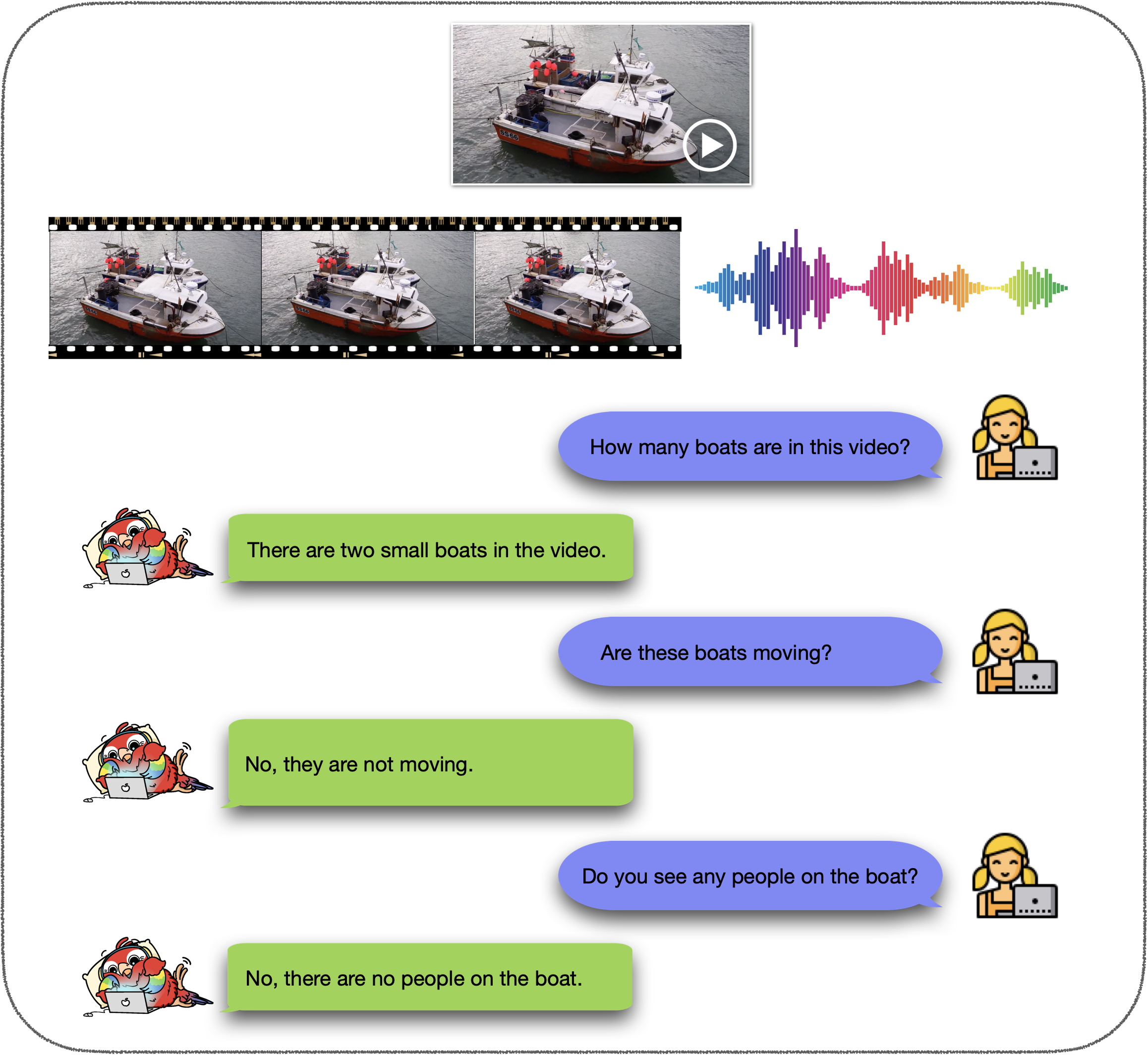}
    \caption{
        An example showcasing \ours's capability in {\bf video-grounded question answering}. \ours can recognize the boats and their amount. Besides, it is able to identify boats' actions over time.
    }
    \label{fig:example_4}
\end{figure}

\begin{figure}[ht]
    \centering
    \includegraphics[width=0.75\textwidth]{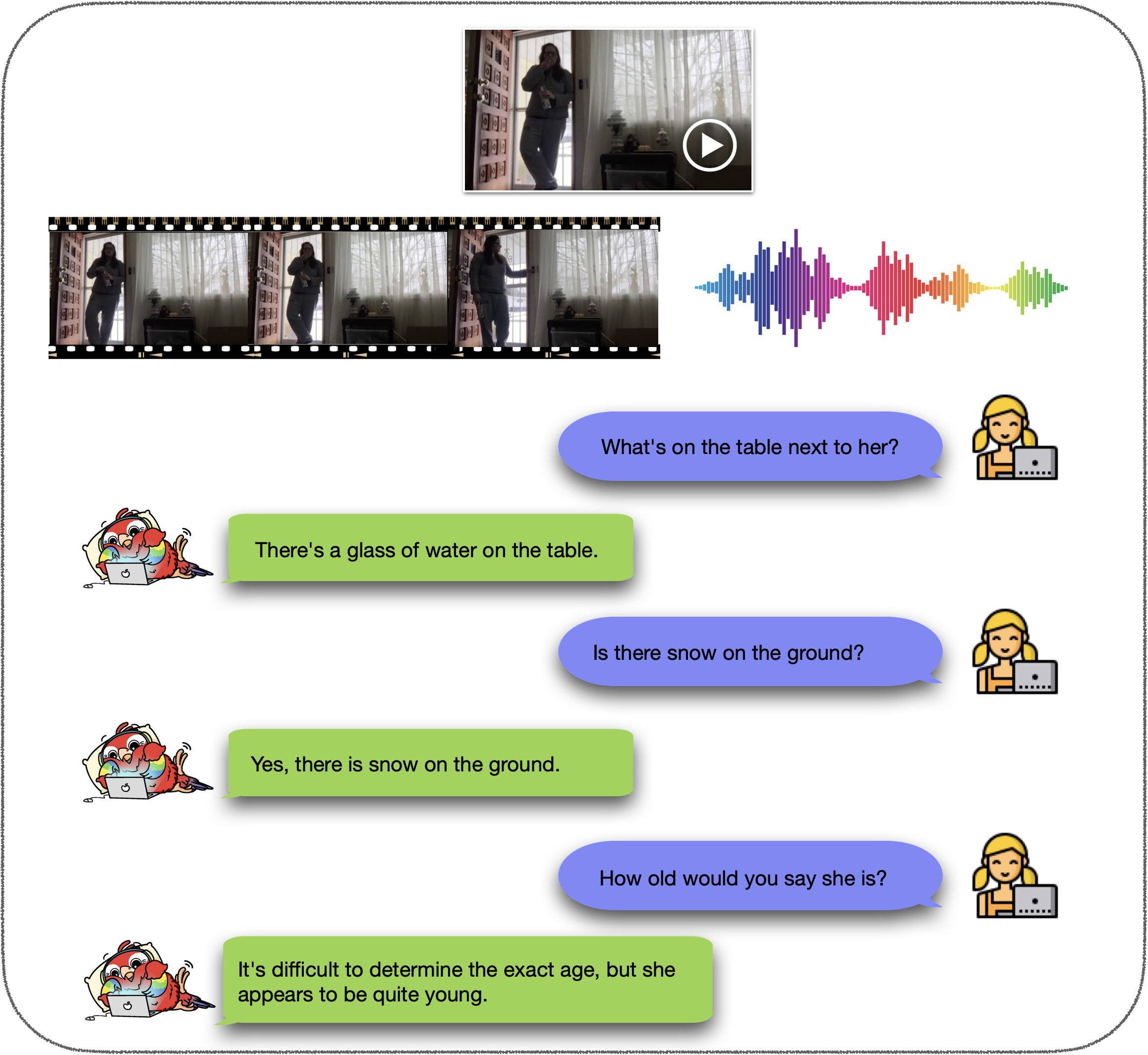}
    \caption{
    An example showcasing \ours's capability in {\bf visual reasoning}. Despite only a small portion of ``white'' being visible outside the door, \ours can infer the presence of ``snow''. Furthermore, it attempts to estimate the age of the woman.
    }
    \label{fig:example_5}
\end{figure}

\begin{figure}[ht]
    \centering
    \includegraphics[width=0.75\textwidth]{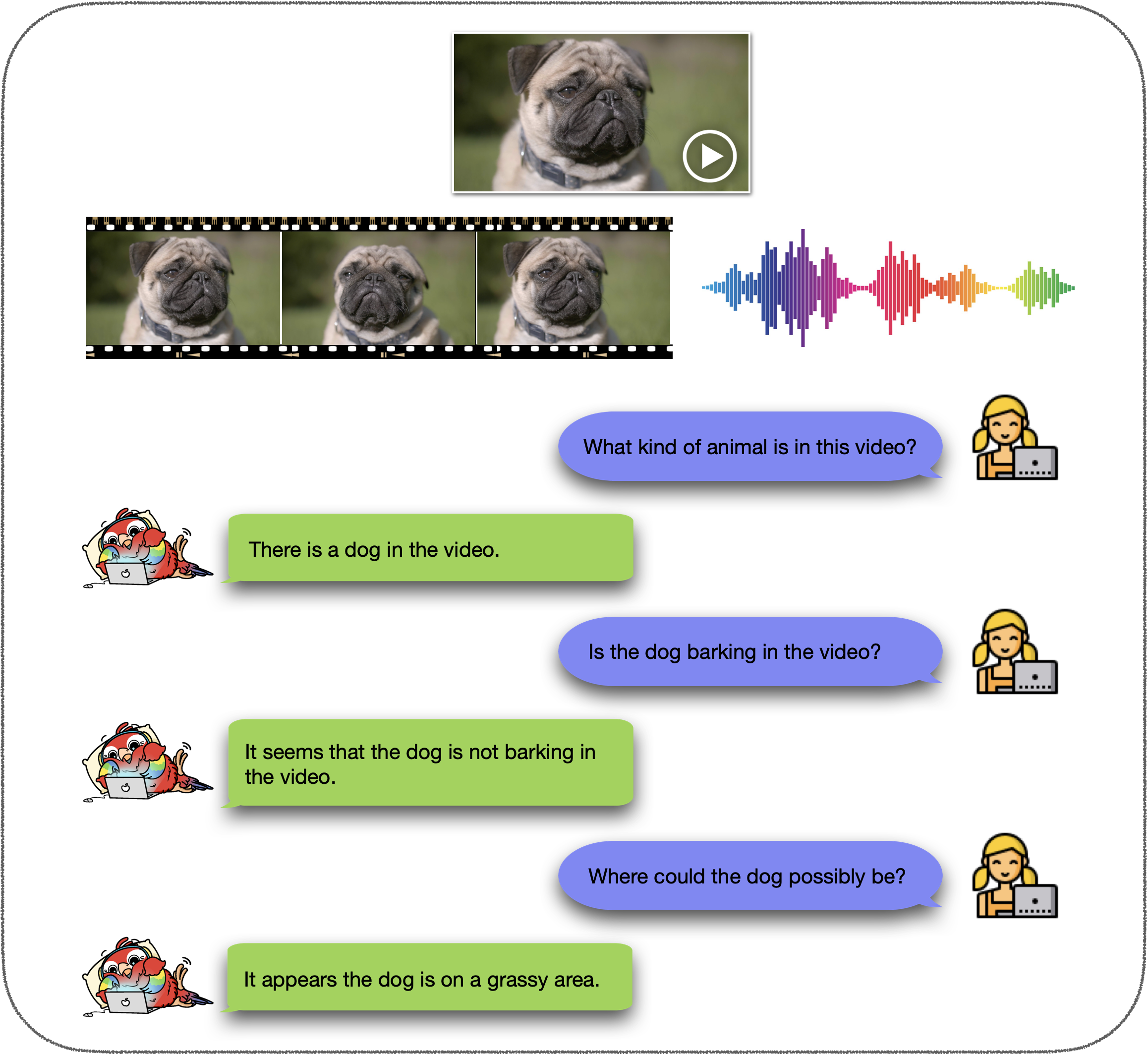}
    \caption{
    An example showcasing \ours's capability in {\bf video- and audio-grounded question answering}. The video showcases a dog on a grassy field, remaining silent as indicated by the audio track.
    }
    \label{fig:example_6}
\end{figure}

We present several examples that highlight the proficiency of our \ours in understanding and following multi-modal instructions.
In \autoref{fig:example_1}, \autoref{fig:example_2}, and \autoref{fig:example_3}, we showcase our system's multi-modal ability to understand and generate responses based on an image. These examples demonstrate how our system comprehends visual content and produces high-quality, fluent responses in natural language conversations. Our system generates contextually relevant and informative answers to various questions about the image, demonstrating its capability to communicate about visual content naturally and fluently.
\autoref{fig:example_4} and \autoref{fig:example_5} present two examples that demonstrate \ours's excellent understanding of videos. We showcase its responses to various questions related to the video content, highlighting its ability to comprehend video information effectively.
Furthermore, \autoref{fig:example_6} demonstrates our system's capacity to process and integrate multiple modalities of information simultaneously. In this example, in addition to answering various video-grounded questions, \ours effectively identifies whether the dog in the video is barking or not.

In summary, the examples provided showcase the impressive capabilities of our system in generating top-notch, contextually appropriate, and logically consistent responses to diverse questions about visual content within a natural language conversation. The proficiency of our system in natural language processing (NLP) and its adeptness in seamlessly incorporating multiple modalities of information underscore its tremendous potential in designing efficient interfaces for human-machine communication.

\clearpage
\section{Limitations}
\label{sec:limitation}
In this section, we summarize the limitations of \ours as follows:
\begin{itemize}[leftmargin=*,topsep=0.1em,itemsep=0.1em,parsep=0.1em]
    \item \textbf{Evaluation}: We show some examples showcasing the multi-modal ability of our \ours. However, we acknowledge that these efforts may not be fully adequate for accurately and comprehensively demonstrate model capabilities. \citet{DBLP:journals/corr/abs-2305-15717} highlights that instruction-tuned LLMs might not perform as well as the reported evaluation results suggest. Hence, we have concerns regarding the ability of our evaluation to provide an accurate reflection of the true capabilities of \ours.
    \item \textbf{Single-Turn Dialogue}: While our training data mainly consists of "dialog-like" instructions, it's important to note that these instructions are currently limited to single-turn interactions. It is crucial to acknowledge that \ours are not currently optimized for handling multi-turn dialogues and may not effectively leverage long-range context.
    \item \textbf{Hallucination, Toxicity and Fairness}: According to empirical evidence presented by \citet{DBLP:journals/corr/abs-2304-14402}, instruction-tuned LLMs may encounter issues such as hallucination, toxicity, and fairness. However, it is important to note that we do not evaluate our models, \ours, in relation to these aspects due to the unavailability of suitable evaluation suites.
\end{itemize}
We acknowledge these limitations and recognize the need for addressing them in future work.

\section{Conclusion and Future Work}

In this paper, we present \ours, a multi-modal instruction-tuned LLM that accommodates four distinct modalities: image, video, audio, and text.
In addition to the standard modality module and cognitive module, we propose a novel approach to align representations from different modality encoders into a shared space. Unlike previous methods, our approach combines representation alignment and instruction tuning into a single step, mitigating potential error propagation during multi-step tuning. Furthermore, we curate \ours instruction dataset, a large-scale dataset of multi-modal instructions using \model{GPT-3.5-Turbo}. We demonstrate examples showcasing the multi-modal understanding ability of \ours.

We discuss the limitations of our work and point out that current multi-modal instruction-tuned LLMs may suffer from various aspects in \autoref{sec:limitation}. We leave the investigation of these issues to the future work. Furthermore, we intend to broaden our corpus to encompass multi-turn and multilingual dialogues. This endeavor will take advantage of the capabilities of LLMs to effectively generate/translate long-document texts~\citep{wang2017exploiting,lyu2023new,wang2023document,wu-etal-2023-document}.

\bibliography{iclr2023_conference}
\bibliographystyle{iclr2023_conference}

\appendix

\end{document}